\newif\ifarxiv
\arxivtrue 

\ifarxiv
	\documentclass[12pt]{article}
	\usepackage[total={6.5in, 9in}]{geometry}
	\usepackage{tgtermes}
	\usepackage[T1]{fontenc}
	\usepackage[urw-garamond]{mathdesign}
\else
	\documentclass[journal]{IEEEtran}
\fi

\usepackage{etoolbox}
\usepackage{ccaption}
\usepackage{amsmath}
\usepackage{amsmath}
\usepackage{graphicx,wrapfig}
\usepackage{amsfonts}
\usepackage{float}
\usepackage{amsthm,amsmath}
\usepackage{enumerate}
\usepackage{graphicx}
\usepackage{url}
\graphicspath{{./fig/}}

\begin{document}  
\ifarxiv
	\author{
		Josh Warren, Jeff Lipkowitz\\
		\textit{University of Chicago}\\
		Vadim O. Sokolov\\
		\textit{George Mason University}\\
	}
	\title{Clusters of Driving Behavior from Observational Smartphone Data}
	\date{First Draft: December 2017\\
		This Draft: January 2018
	}
	\maketitle
\else
	\title{Clusters of Driving Behavior from Observational Smartphone Data}
	\author{Josh Warren, Jeff Lipkowitz, and Vadim Sokolov
	\thanks{J. Warren and J. Lipkowitz  are with the University of Chicago, Chicago, IL, 60637, USA  e-mail: \{warrenjosh,jlipkowitz\}@uchicago.edu.}
	\thanks{V. Sokolov  is with the George Mason University, Fairfax, VA, 22030  e-mail: vsokolov@gmu.edu.}
}
	\IEEEpeerreviewmaketitle	
\fi

\begin{abstract}
Understanding driving behaviors is essential for improving safety and mobility of our transportation systems. Data is usually collected via simulator-based studies  or naturalistic driving studies. Those techniques allow for understanding relations between demographics, road conditions and safety. On the other hand, they are very costly and time consuming. Thanks to the ubiquity of smartphones, we have an opportunity to substantially complement more traditional data collection techniques with data extracted from phone sensors, such as GPS, accelerometer gyroscope and camera. We developed statistical models that provided  insight into driver behavior in the San Francisco metro area based on tens of thousands of driver logs.  We used novel data sources to support our work. We used cell phone sensor data drawn from five hundred drivers in San Francisco to understand the speed of traffic across the city as well as the maneuvers of drivers in different areas. Specifically, we clustered drivers based on their driving behavior. We looked at driver norms by street and flagged driving behaviors that deviated from the norm.
\end{abstract}

\section{Introduction}
\ifarxiv
Understanding
\else
\IEEEPARstart{U}{nderstanding} 
\fi
driving behavior is of increasing importance in the context of aging population and increasing vehicle automation. Novel vehicle technologies and mobility paradigms change the driving habits quicker than traditional data sources such as naturalistic driving studies, are able to reflect. A specific example is vehicle automation, which allows for drivers to stay disengaged for long periods of time. The increased level of automation changes the driving patterns. The federal government, transit agencies, insurance companies, fleet managers and family members of young or elderly drivers are interested in understanding those changes in driving behaviors which is challenging when  out-dated information and methodologies are used. 

Historically, driving simulators and naturalistic driving studies are the main sources of data on driving behaviors. Simulator based studies provide tight control of driving contexts and allow for repetitive experiments; thus they provide a better data set from a classical statistical modeling point of view. Naturalistic studies, on the other hand, allow for analyzing behaviors in a large number of scenarios and do not assume any constrains on the driving situations experienced by the participants. Both approaches have two major drawbacks. First is the cost of a study. A typical research study involves at the most several dozen participants. Second, it takes several years to set up a study to collect the data and analyze it. This long time frame means that data collected and analyzed might get outdated due to changing vehicle technologies or demographics. The data collection of the SHRP2 naturalistic driving study was completed in early 2014. The data collection effort cost several million dollars and many of the analysis projects based on this data are currently ongoing. Though, this driving study the largest and the most comprehensive NDS ever undertaken, it only involved 2141 vehicles in six urban areas around the country \cite{campbell2012shrp}. 

At the same time, drivers generate data while using smartphone apps dedicated to collecting data from vehicles and its surroundings. Continuous data collection using unobtrusive technology, such as a smartphone, provides a viable alternative to study-based data collection. There are several popular applications available for smartphones that are used by a large number of users and collect very detailed data about driving behaviors. Data from those applications addresses both of the issues mentioned about the simulator-based or naturalistic studies. Data is continuously collected from a very large number of users, e.g. millions of people.  Thus, the main advantage of using observational data from smartphones are (i) orders of magnitude larger sample size, (ii) continuous data collections and the absence of the delay between formulating the research question to obtaining analysis results. While data collection from smartphones is common, there are no existing efforts that demonstrate statistical learning models that can analyze large-scale observational data sets. 

In this paper we develop an analytical ``pipeline'' to annotate driving behaviors and to identify disciplined and abnormal driving behaviours. Our approach is unsupervised and does not rely on any human-labeled samples. We rely on a large amount of data available for many road segments in San Francisco to understand the distribution of driving behaviours and then identify five distinct clusters of drivers. 

The contribution of this paper lie in four aspects:
\begin{itemize}
\item We implement end-to-end processing and a statistical learning pipeline from the raw observations from smartphones to the distribution of driving behaviours, conditional on the time of the day and the day of the week
\item  To the best of the authors' knowledge, this is the first work using an unsupervised approach to identify distinct driving behaviours using clustering algorithms from the smartphone data
\item We demonstrate our methodology using a sample of twenty one thousand trips in San Francisco
\item A distributed implementation of the statistical learning algorithms using Microsoft's cloud platform Azure. 
\end{itemize}

\section{Related Work}
Driving behavior analysis is an interdisciplinary field and has drawn increasing attention in the recent decade, specifically in the context of vehicle automation \cite{ohn-bar2016}. An in-depth recent review of the literature on driving style analysis frameworks is given in \cite{sagberg2015}.

Traditionally, the main source of the data for the driving behavior analysis are naturalistic driving studies (NDS). A usual NDS involves instrumenting a vehicle with data collection equipment, such as video cameras and GPS. NDS participants usually do not have any specific instructions and thus, the data collected provides an unbiased view of how people actually drive in everyday life. A large scale NDS has been recently completed by the National Highway Traffic Safety administration and Virginia Department of Transportation \cite{dingus2006}. Hundred vehicle participated in the study and data on proximately 2 million vehicle miles was collected. One of the largest NDS was performed under the second Strategic Highway Research Program (SHRP 2) \cite{campbell2012shrp}. One of the analysis efforts based on this data studies the relationship between anger and near-crash or crash risk \cite{precht2017}. That analysis showed that anger results in dangerous and aggressive behavior. The anger state was estimated based on the reactions of the drivers to  driving errors. Another paper \cite{guo2010} analyzes the 100-car NDS for understanding the relations between demographic and personal characteristics with the driving behavior. Due to small number of crashes observed in an NDS, several surrogate events, such as near-crasher are also used to understand the safety metrics \cite{guo2010}. A framework for analyzing impacts of vehicle automation on driving behavior was presented in \cite{ohn-bar2016}. The paper presents a modeling framework that incorporates driver's reactions to events inside and outside the vehicle. A methodology to measure the aggressiveness of driving behavior is presented in \cite{feng2017}.

Data collected from smartphone provides a viable alternative to naturalistic driving studies \cite{castignani2015,goncalves2014,paefgen2012}. Smartphone sensors typically have larger errors compared to dedicated instruments used for NDS. A recent review of using a smartphone for driving data collection is presented in \cite{engelbrecht2015}. Machine learning techniques allow for extraction of several types of events from cell phone data. Discriminative machine learning algorithms were developed in \cite{wallace2014}. Data from 4 drivers from Candrive study was used to create a metric, called driving signature. This metric is used to distinguish one driver from another based on velocity, acceleration and route choices.  In \cite{cruz-silva2013} authors used data mining techniques to extract 159 different features, from smartphone data, that can be used to distinguish one driver from another.  An in-depth review of machine learning and artificial intelligence techniques for analyzing driving behaviours is presented in \cite{meiring2015}. Another set of machine learning techniques was developed in \cite{hong2014} to analyze and predict aggressive driving behaviours. An advanced Dynamic Time Warping (FastDTW) was used in \cite{liu2017a} to recognize the vehicle steering based on the accelerometer and gyroscope measurements. An algorithms for estimating acceleration by fusing GPS and accelerometer measurement was proposed in \cite{yu2016,chowdhury2014}. An analysis similar to ours was performed by \cite{wallace2014}. Authors used data from 14 older drivers to find similarities between pairs of drivers. Data from Norway was used to identify driving clusters \cite{nordfjaern2013}. Four distinct clusters of drivers were  identified based on the survey of 1731 drivers. Older drivers are of particular interest. Different slowly progressing diseases such as Alzheimer's can lead to significant decrease in driving abilities. An predictive model for identifying potential drivers with dementia was developed in \cite{eby2012}. A detailed analysis of deceleration behavior based on data from one driver was presented in \cite{wallace2015}. Data from Candrive project was analyzed and two district phases of deceleration (before 27.5 km/h and after) were were identified. In smartphone-based studies, it is also typical to combine phone sensor data with the data from  Controller Area Network Bus (CAN-BUS) \cite{carmona2015}. The CAN data is retrieved via the On-Board Diagnostics (OBDII) interface, which has been available on all of the new vehicles sold in the US since 1996. Another advantage of smartphone includes availability of significant computing power and Internet connection. Thus, it makes it possible to use smartphones for real-time applications \cite{bergasa2014}.

\section{Data Pre-Processing}
In this work we are interested in using data collected from smartphone sensors to understand the district driving behaviours among the San Francisco drivers. Privacy considerations and lack of ground truth data prevent us from using supervised learning approaches. Thus, the unsupervised learning is the required approach for our data set. Our data was provided by an  internet of things (IOT)  startup company called Nexar. This company develops an app used by tens of thousands of drivers in the US to record data from phone sensors, including GPS, accelerometer, gyroscope and camera used as a dashcam. The data is continuously transferred from the phone to the company via mobile connection and is logged by the company on their data servers. The main offering is an AI dash-cam mesh network that will alert drivers in real-time with warning to dangers on the road based on the camera's vision and community alerts from other drivers. The app provides alerts of upcoming traffic changes and urgent warnings to drivers to help them avoid accidents.

A data set of rides from June 2016 was provided for analysis. Company's software stores ``rides'' which are  driver sessions with the app running in the background and don't necessarily correlate to a single ride. Data for each ride is broken down into different sensor files. GPS, accelerometer, magnetometer, CoreMotion and gyroscope sensor measurements were provided by ride with varying measurement frequencies. GPS and magnetometer are measured at 1Hz while accelerometer, gyroscope, and CoreMotion measurements are collected at 10Hz.

\paragraph{\textbf{Speed and Acceleration}} To address the issue of noisy accelerometer and GPS-measured speed data, several filtering techniques were applied. After comparing the results from the $\ell_1$-trend filter \cite{kim2009}, total variational denoisising \cite{rudin1992nonlinear}, and the moving average filter \cite{hyndman2014forecasting}, we have chosen the latter one. Moving average filters with 1 second time window were applied to the accelerometer and gyroscope data. Moving average filters  were also suggested in the previous work \cite{cruz-silva2013,liu2017a}. The processing for each sensor varied, and we will briefly describe some of the transformations needed to make use of them. 

\begin{figure}[H]
\includegraphics[width=1\linewidth]{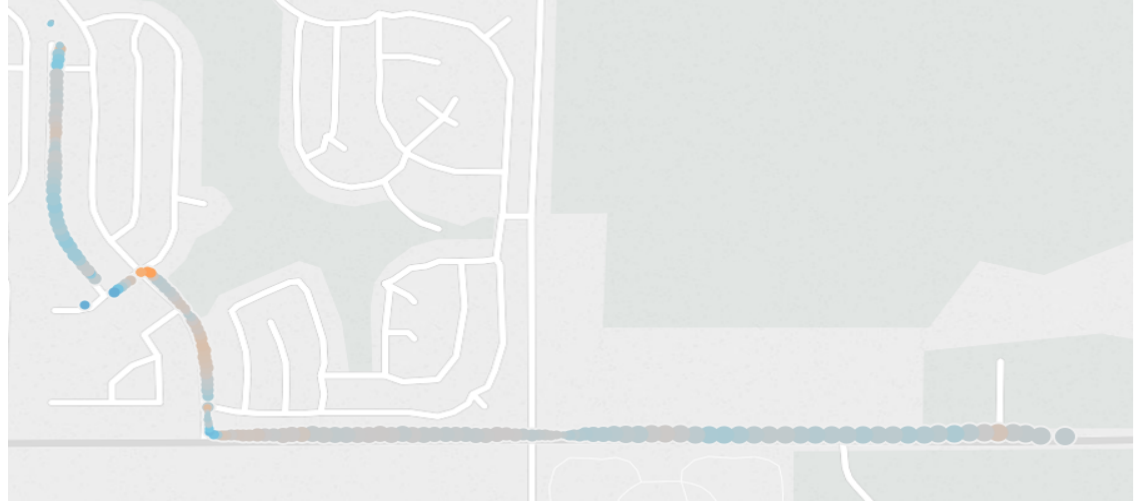}
\caption{Y-component of the acceleration signal. Gray is modest acceleration, blue is modest deceleration. Orange is rapid deceleration. }
\label{fig:accel-map}
\end{figure}

The accelerometer measures three directions of force along the X, Y, and Z-axes. Acceleration
values for each axis are reported directly by the hardware as G-force values. Therefore, a value of 1.0 represents a load of approximately 1-gravity (Earth's gravity). The Y-axis goes along the tall side of a screen and the Z-axis goes ``through" the phone perpendicular to the screen surface. Thus, when the phone is placed on dashboard a with camera facing the windshield, the Z-direction corresponds to the longitudinal movement of the vehicle. To understand which axes corresponds to the lateral movement of a car, we need to understand the phone's rotation which was not available. Thus, we the took second most dominating signal near road intersections and treated it as lateral movement axes. For example, Figure \ref{fig:accel-map} shows Y-axes plotted on the map.

It is clear that Y was capturing the side-to-side maneuvers of the vehicle. This whole transformation in the interpretation of sensors was the result of putting the phone in landscape mode, which seemed to be the primary way drivers used the app. This transformation seemed to be consistent across the dataset.

Speed was a much cleaner variable. It is measured every second by the GPS and produces much smoother changes overtime. The speed data closely aligned with real world expectations. Faster speeds were found on highways with slower speed mostly found on local roads.

\paragraph{\textbf{Location Data}} Our unit of analysis was a road segment and the goal is to understand the distribution of driving behaviours for each of the road segments for which enough data is collected to make statistically rigorous conclusions.
While phone GPS location measurements are good enough for most daily tasks, they are not accurate enough for quality road positioning. A phone's latitude and longitude measurements are routinely found in unexpected locations.
\begin{figure}
	\includegraphics[width=1\linewidth]{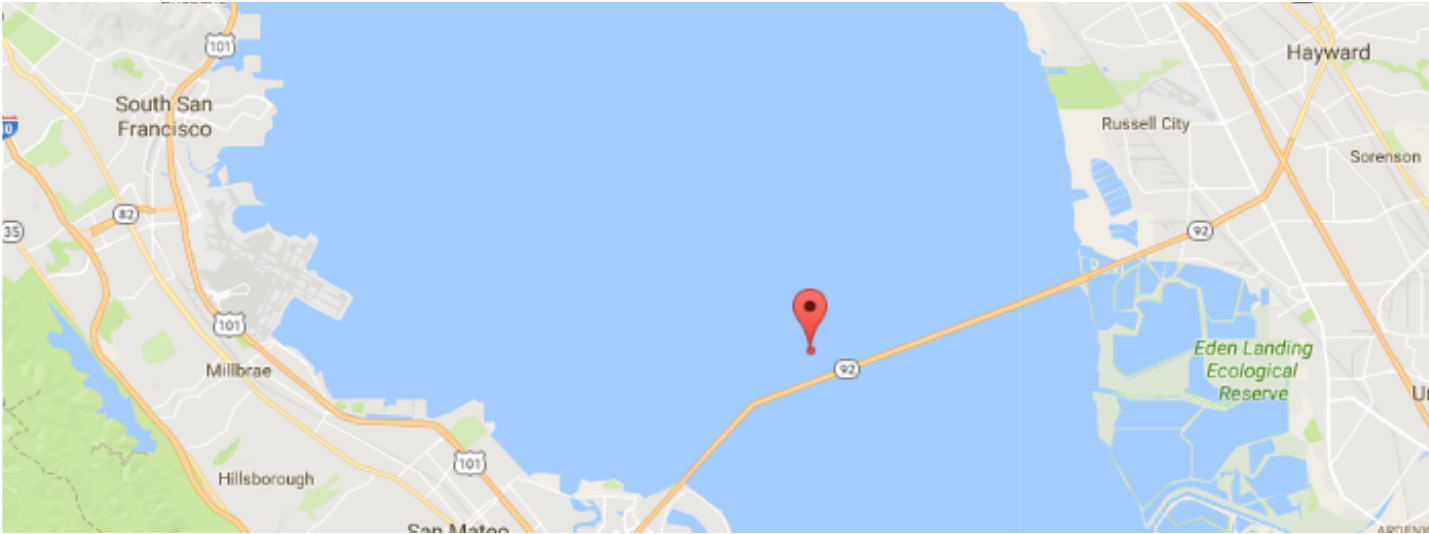}
	\caption{Incorrect GPS location measurement.}
\end{figure}

We had to address the problem of map matching \cite{assessing_energy_trb1}. The goal of map-matching is to infer a map feature (point or line) associated with a sequence of GPS measurements. The measurement error is the error in estimated coordinates and associated time step. The error in positional GPS measurement can be described by a bivariate normal distribution \cite{leick2015gps}. The typical error in augmented GPS measurements is in the range between eight and two meters \cite{bergasa2014}. The sampling error introduces uncertainty of vehicles position in between the measurements. The task of assigning a single GPS point (as in navigation applications) or a sequence of GPS points (as in post-processing applications used in travel modeling) to a set of map features is called the map matching problem. The goal is to infer the actual location on the map, that the traveler visited based on the noisy latitude, longitude and possibly speed measurements from GPS, and in the case of post-processing, to have the trajectory resolve into a consistent route through the transportation network. Both measurement error and infrequent sampling rate require attention while developing map-matching algorithms. We used Google Maps Roads API \cite{googleapi}, which allows us to access map-matching function. This API takes a set of up to 100 GPS values and returns the most likely points that correspond to each reading. This helps correct off-road points as well as points that improbably dart around intersections jumping from one road segment to another. The accelerometer measurements notorious for having high levels of noise. Our resampling method to generate measurements each second instead of 10 times a second greatly reduced the noise to  a satisfactory level  for further analysis. 

In an Azure cloud machine, we processed 20,760 log files and made 30M + Google API calls, resulting in a 5TB database of over 10M unique latitude and longitude pairs. Given the amount of data, we had a large number of samples for almost every major road and freeway in San Francisco. Further, for many road segments we had enough samples (more then 40 trips) for each 30 minute interval during rush hour and were able to model the distributions conditional on the time of the day. 

\section{Empirical Results}
\begin{figure}[H]
	\centering
	\includegraphics[width=\linewidth]{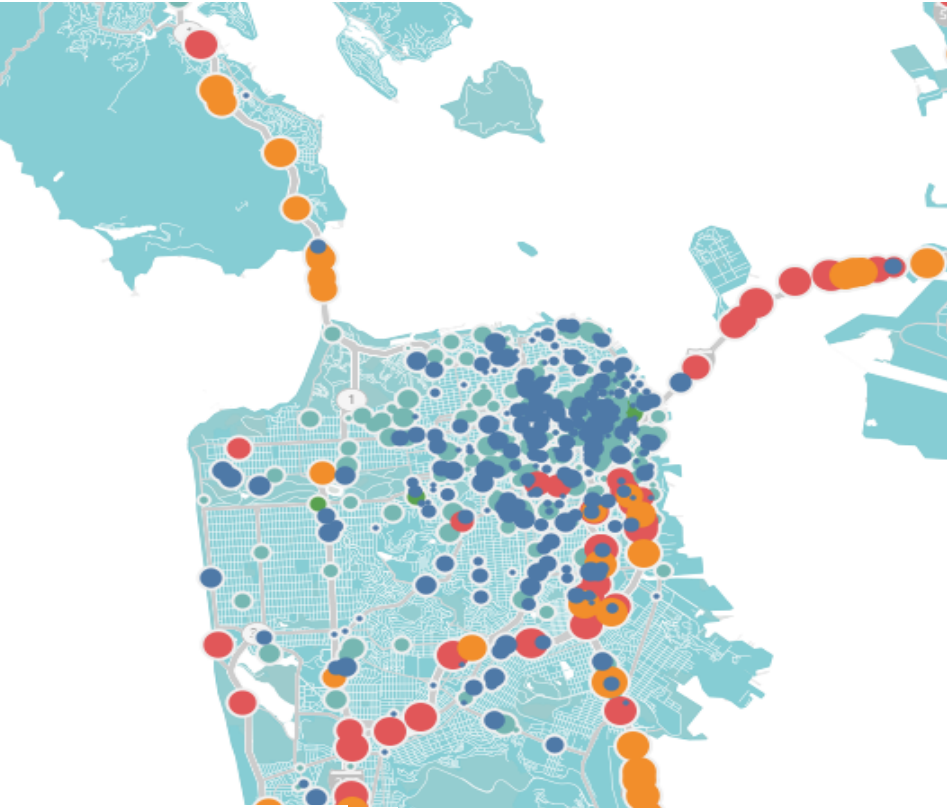}
	\caption{Clustering of driving behaviors on individual road segments. Each road segment is colored according the cluster it belongs to, i.e all road with blue circle belong to the same cluster.}
	\label{fig:map-speed}
\end{figure}

With processed data in place we begin to evaluate if our data would be complete enough to produce good street level norms that could be used to flag abnormal driving behaviors. We looked at identifying the type of road solely on the basis of our sensor data. Figure \ref{fig:map-speed} shows the result of our first unsupervised clustering algorithms. Here, we can see that highways are covered with red and orange dots while the dots around downtown San Francisco are colored blue and green. Our unsupervised learning algorithm can clearly distinguish highway driving from city driving.
Figure \ref{fig:map-speed}.

Next, we find the distributions over the driving behavior parameters for individual road segments and use those to distinguish ``good'' drivers from aggressive ones. Using this data, we then clustered the drivers to uncover group trends.

\begin{figure}[H]
	\centering
	\includegraphics[width=\linewidth]{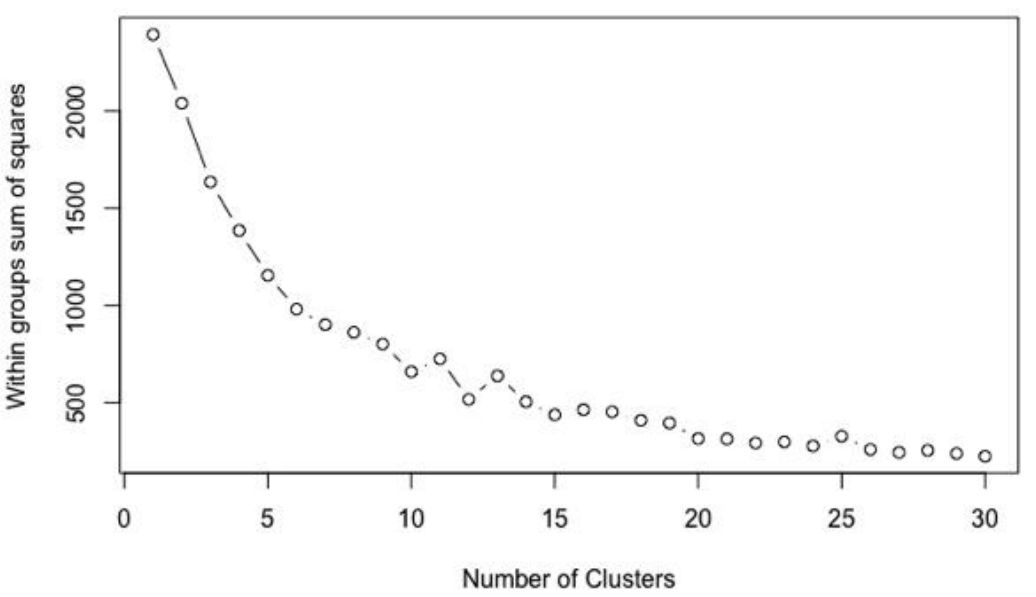}
	\caption{Dependence on variance reduction from number of clusters.}
	\label{fig:elpbow}
\end{figure}

\begin{figure}[H]
	\centering
\includegraphics[width=\linewidth]{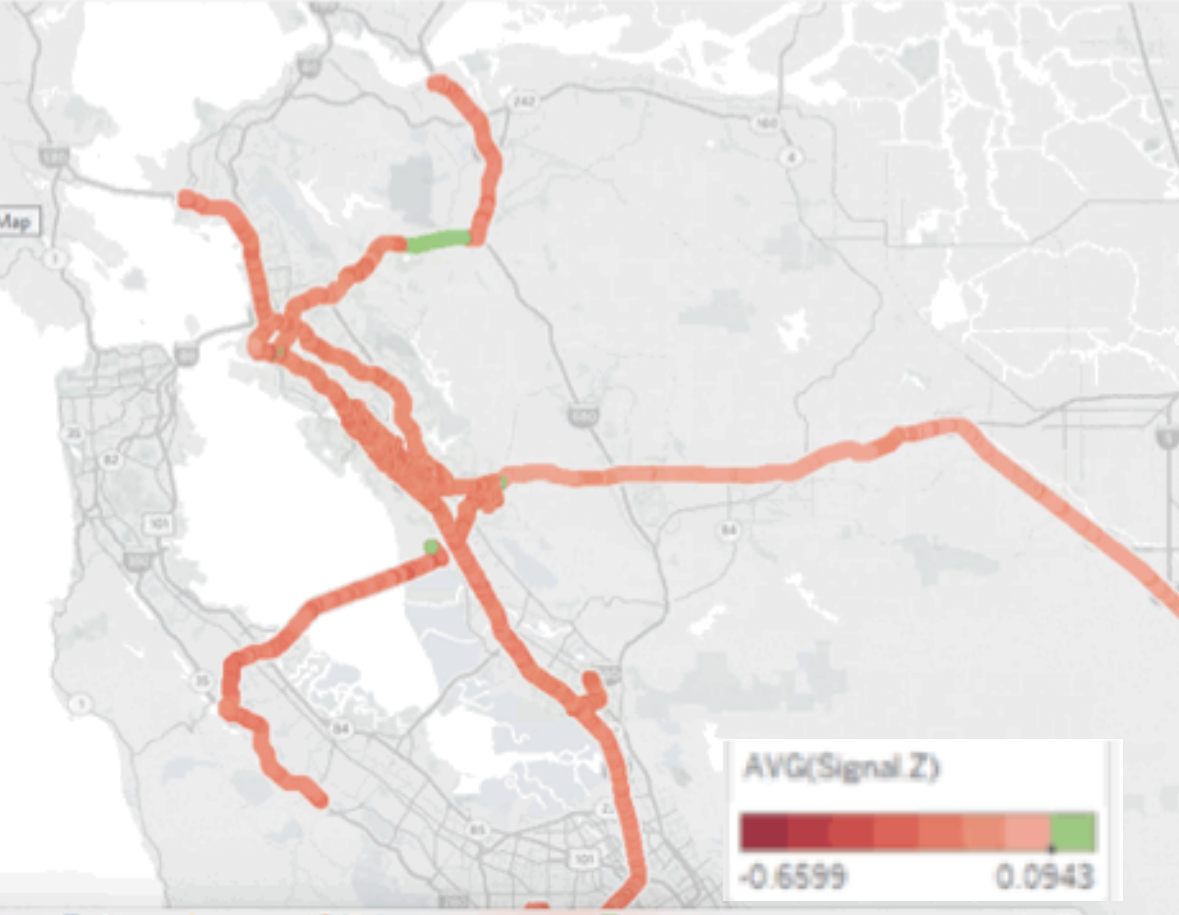}
\caption{Selected acceleration trajectories of drivers from cluster 4, who tend to drive smoother and slower then average. Red color corresponds to smooth acceleration.}
\label{fig:cluster-4}
\end{figure}

We looked at the within group sum of squares, as shown in Figure \ref{fig:elpbow}  to understand the optimal number of groups. 
We initially used six clusters, but found that five actually produced more distinct clusters. Users were broken into five distinct clusters based on our classification of the users' behaviors on the road. Additional details like age, gender, and past incidences would be helpful for clustering, but due to privacy, those were unavailable. Originally we thought that there would be a group of users who would be classified as consistent speeding offenders and would simply flag the user as outside the norm based on our classification. However this group was not flagged during our clustering. There was no "lead foot cluster" when we normally think of traditional drivers. We found drivers are more readily classified by their general driving maneuvers picked up by the accelerometer and gyroscope \cite{liu2017a}. The largest cluster in our data set were users that by in large conformed to road norms, not accelerating quickly or taking sharp turns.
The drivers in Cluster 4 tended to drive slower than the norm. Knowing this we also observed that users with this group also had a behavior pattern associated with rapid breaking. Users in this group preferred to ride slow and were conservative with their surroundings. Further analysis would have to be done to understand what was the root cause of these behaviors and relationships. Did users in this group have issues with their car? Is this traffic related?
Figure \ref{fig:cluster-4} shows a selection of rides from cluster 4 drivers - they tended to be slower than average.

Cluster 3 was the opposite extreme of cluster 4. These drivers were very aggressive. As we mentioned earlier, there was no large "Speeding class", but Cluster 3 came close. Drivers in this group were very quick to accelerate as well as decelerate. This kind of jerky motion can be linked to higher risk of accidents \cite{karapiperis2015usage}. These drivers are more likely to aggressively tailgate and accelerate too suddenly. For the most part these drivers stayed within the confines of the speed limit. However, they were very aggressive in maximizing their speed to the limit. We see this by the aggressive changes in both X and Y components of the accelerometer measurements. 


\begin{figure}[H]
\begin{tabular}{cc}
\includegraphics[width=0.5\linewidth]{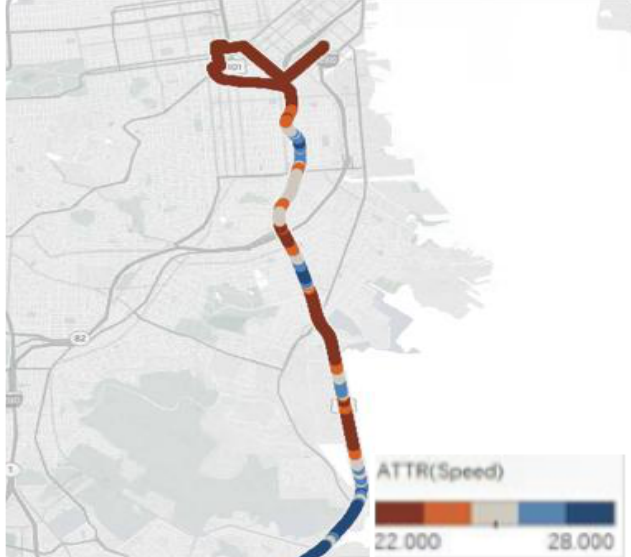} & \includegraphics[width=0.5\linewidth]{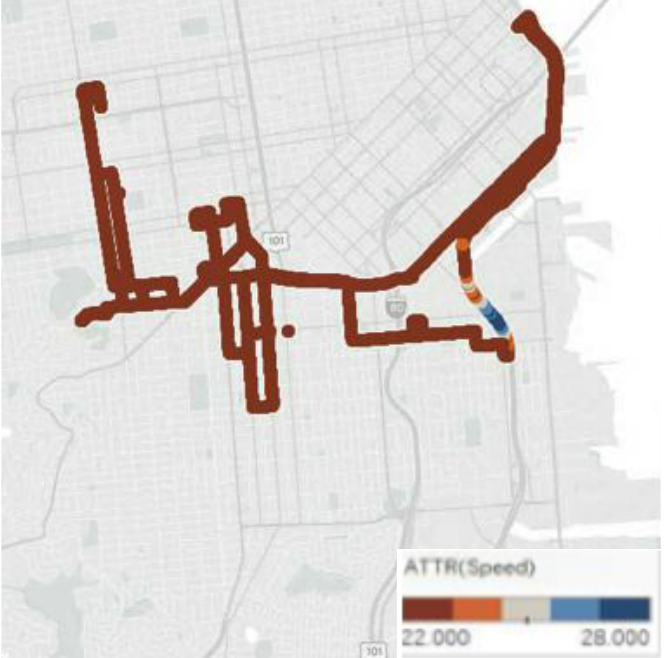}\\
(a) Aggressive behavior & (b) Not aggressive behavior
\end{tabular}
\caption{Two examples of individual speed trajectories for drivers from cluster 3.}
\label{fig:cluster-3}
\end{figure}

Figure \ref{fig:cluster-3} shows individual rides from cluster 3. The aggressive driving behavior is not present in every trip for drivers from cluster 3 (panel (b) on Figure \ref{fig:cluster-3}). Some trips were closer to the trips from the cluster 4 (the baseline). However, when observed on another trip the same driver (panel (a) on Figure \ref{fig:cluster-3}) does drive aggressively. This change in the level of aggressiveness can possibly be explained by the surrounding traffic conditions, e.g. in dense traffic a driver does not have enough space for rapid acceleration/deceleration. This shift in behavior requires further analysis.

The drivers in Cluster 2 tended to decelerate faster than normal. It's unclear what was the root cause of this behavior. One possibility is that they were closely tailgating other drivers in bumper-to-bumper traffic. Another differentiating factor of Cluster 2 is the way they turn, we saw a lot more extreme lateral accelerometer forces from them indicating they turn sharply.

\begin{figure}[H]
	\includegraphics[width=1\linewidth]{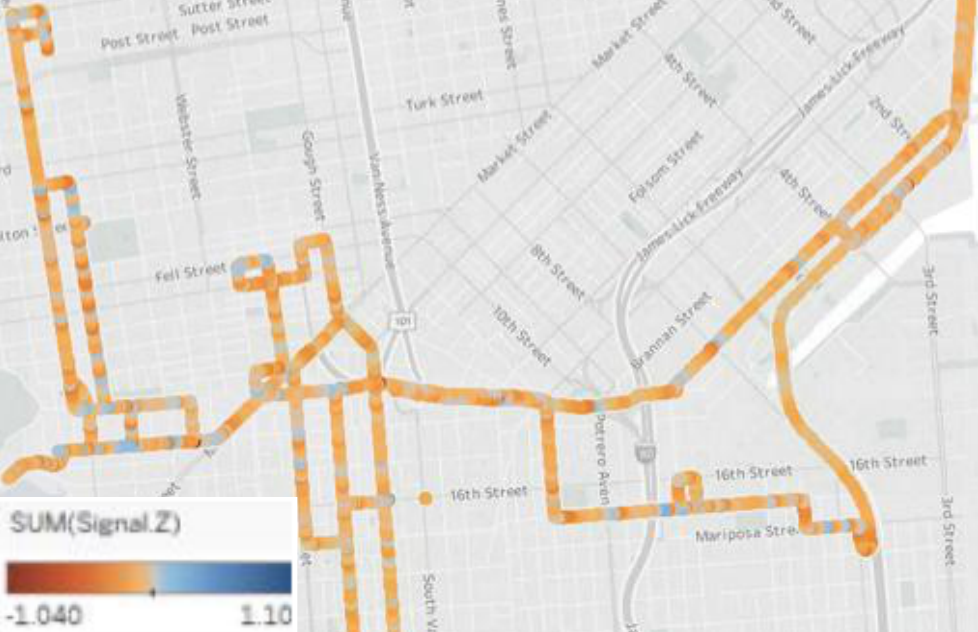}
	\caption{Longitudinal acceleration of a driver from cluster 2.}
	\label{fig:15}
\end{figure}

Let's review an individual rider from cluster 2, focusing on the Y accelerometer values. We observed that when users make left turns they tend to stay in the red range while users turning in the right direction tended stay in blue color range. This is an indication of sharp turns in both directions. Figure \ref{fig:15} shows the Z value of the accelerator.

This behavior shows that the user  constantly used his breaks. One possible explanation is that  users are in heavy traffic and require constant tap and go.
Looking further at the user's speed, we see that the hypothesis is supported based on the behavior. The average speed in San Francisco is 25 mph and the users never never rose  above 22 mph. This is a clear indication that traffic condition were heavy. It is  also important to note that this cluster has extremely similar behaviors to that we see is cluster 4. However, the major difference is tight turns in the Y dimension. If we were change the number cluster from 5 down to four, then these groups would merge into one.

\section{Discussion}\label{discussion}
We demonstrated how to use unsupervised learning to identify distinct driving patterns from a large scale observational data set collected form users' smartphones. Specifically, we used  accelerometer, GPS location and gyroscope measurements to create a user profile. We were able to create the ranking profile of each user by identifying the norms of the population identifying scenarios that each user didn't confirm the norm. After profiling all of the users we performed cluster analysis to identify various driver groups. We found that five cluster tended to be the best way to cluster the population

After examining each group, we found user behavior in the city of San Francisco can be classified into distinct groups such as aggressive drivers, conservative drivers and conservative drivers who make aggressive turns. 

There are many directions for future research. The same smartphone app is used to record videos that are further analyzed by the company to identify crash and near-crash events. Understanding the relationship between everyday driving styles and proneness to accidents will lead to better understanding of the possible safety implications of aggressive driving. The data set collected from smartphone users suffers from a bias issue. Most of the users are ride share drivers, e.g. Lyft and Uber. Understanding the  demographics of the users on an aggregate level will allow for the development of statistical methodologies to correct for  bias. As we mentioned earlier, some of the driving behaviors can be explained away by traffic conditions. Changing statistical learning models so that traffic is accounted for will allow to further investigate the relations between traffic, driving behavior and safety. Lastly, the data set collected from app users provide a large but ``shallow'' sample. We do not have any sociodemographic characteristics about the drivers. To better understand the relations between socio-demographic characteristics,  driving experience and driving behavior, some data fusion methodologies need to be developed that combine data from naturalistic driving studies and observational data from smartphones.

\section*{Acknowledgments} 
We are grateful for data and support provided by Nexar. A grant from Microsoft for providing the storage, computing time and software  necessary to perform the analysis. We thank Sema Barlas for many insightful suggestions on the design and presentation of the analysis.

\bibliography{ref}

\begin{thebibliography}{10}

\bibitem{googleapi}
{Snap to Roads | Google Maps Roads API }.
\newblock \url{https://developers.google.com/maps/documentation/roads/snap}.
\newblock Accessed: 2017-07-10.

\bibitem{bergasa2014}
L.~M. Bergasa, D.~Almer{\'\i}a, J.~Almaz{\'a}n, J.~J. Yebes, and R.~Arroyo.
\newblock {{DriveSafe}}: {{An}} app for alerting inattentive drivers and
  scoring driving behaviors.
\newblock In {\em 2014 {{IEEE Intelligent Vehicles Symposium Proceedings}}},
  pages 240--245, June 2014.

\bibitem{campbell2012shrp}
Kenneth~L Campbell.
\newblock The {{SHRP}} 2 naturalistic driving study: {{Addressing}} driver
  performance and behavior in traffic safety.
\newblock {\em TR News}, (282), 2012.

\bibitem{carmona2015}
Juan Carmona, Fernando Garc{\'\i}a, David Mart{\'\i}n, Arturo de~la Escalera,
  and Jos{\'e}~Mar{\'\i}a Armingol.
\newblock Data {{Fusion}} for {{Driver Behaviour Analysis}}.
\newblock {\em Sensors}, 15(10):25968--25991, October 2015.

\bibitem{castignani2015}
G.~Castignani, T.~Derrmann, R.~Frank, and T.~Engel.
\newblock Driver {{Behavior Profiling Using Smartphones}}: {{A Low}}-{{Cost
  Platform}} for {{Driver Monitoring}}.
\newblock {\em IEEE Intelligent Transportation Systems Magazine}, 7(1):91--102,
  Spring 2015.

\bibitem{chowdhury2014}
A.~Chowdhury, T.~Chakravarty, and P.~Balamuralidhar.
\newblock Estimating true speed of moving vehicle using smartphone-based
  {{GPS}} measurement.
\newblock In {\em 2014 {{IEEE International Conference}} on {{Systems}},
  {{Man}}, and {{Cybernetics}} ({{SMC}})}, pages 3348--3353, October 2014.

\bibitem{cruz-silva2013}
Nuno Cruz-Silva, Jo{\~a}o Mendes-Moreira, and Paulo Menezes.
\newblock Features {{Selection}} for {{Human Activity Recognition}} with
  {{iPhone Inertial Sensors}}.
\newblock 2013.

\bibitem{dingus2006}
T.~A. Dingus, S.~G. Klauer, V.~L. Neale, A.~Petersen, S.~E. Lee, J.~D.
  Sudweeks, M.~A. Perez, J.~Hankey, D.~J. Ramsey, S.~Gupta, C.~Bucher, Z.~R.
  Doerzaph, J.~Jermeland, and R.~R. Knipling.
\newblock The 100-{{Car Naturalistic Driving Study}}, {{Phase II}} -
  {{Results}} of the 100-{{Car Field Experiment}}.
\newblock April 2006.

\bibitem{eby2012}
David~W. Eby, Nina~M. Silverstein, Lisa~J. Molnar, David LeBlanc, and Geri
  Adler.
\newblock Driving behaviors in early stage dementia: {{A}} study using
  in-vehicle technology.
\newblock {\em Accident Analysis \& Prevention}, 49:330--337, November 2012.

\bibitem{engelbrecht2015}
Jarret Engelbrecht, Marthinus~Johannes Booysen, Gert-Jan van Rooyen, and
  Frederick~Johannes Bruwer.
\newblock Survey of smartphone-based sensing in vehicles for intelligent
  transportation system applications.
\newblock {\em IET Intelligent Transport Systems}, 9(10):924--935, December
  2015.

\bibitem{feng2017}
Fred Feng, Shan Bao, James~R. Sayer, Carol Flannagan, Michael Manser, and
  Robert Wunderlich.
\newblock Can vehicle longitudinal jerk be used to identify aggressive drivers?
  {{An}} examination using naturalistic driving data.
\newblock {\em Accident Analysis \& Prevention}, 104:125--136, July 2017.

\bibitem{goncalves2014}
J.~Gon{\c c}alves, J.~S.~V. Gon{\c c}alves, R.~J.~F. Rossetti, and
  C.~Olaverri-Monreal.
\newblock Smartphone sensor platform to study traffic conditions and assess
  driving performance.
\newblock In {\em 17th {{International IEEE Conference}} on {{Intelligent
  Transportation Systems}} ({{ITSC}})}, pages 2596--2601, October 2014.

\bibitem{guo2010}
Feng Guo, Sheila Klauer, Jonathan Hankey, and Thomas Dingus.
\newblock Near {{Crashes}} as {{Crash Surrogate}} for {{Naturalistic Driving
  Studies}}.
\newblock {\em Transportation Research Record: Journal of the Transportation
  Research Board}, 2147:66--74, October 2010.

\bibitem{hong2014}
Jin-Hyuk Hong, Ben Margines, and Anind~K. Dey.
\newblock A {{Smartphone}}-based {{Sensing Platform}} to {{Model Aggressive
  Driving Behaviors}}.
\newblock In {\em Proceedings of the {{32Nd Annual ACM Conference}} on {{Human
  Factors}} in {{Computing Systems}}}, CHI '14, pages 4047--4056, New York, NY,
  USA, 2014. {ACM}.

\bibitem{hyndman2014forecasting}
Rob~J Hyndman and George Athanasopoulos.
\newblock {\em Forecasting: principles and practice}.
\newblock OTexts, 2014.

\bibitem{karapiperis2015usage}
D~Karapiperis, A~Obersteadt, A~Brandenburg, S~Castagna, B~Birnbaum,
  A~Greenberg, and R~Harbage.
\newblock Usage-based insurance and vehicle telematics: insurance market and
  regulatory implications.
\newblock {\em CIPR Study Series}, 1:1--79, 2015.

\bibitem{kim2009}
Seung-Jean Kim, Kwangmoo Koh, Stephen Boyd, and Dimitry Gorinevsky.
\newblock {$\ell_1$} {{Trend Filtering}}.
\newblock {\em SIAM review}, 51(2):339--360, 2009.

\bibitem{leick2015gps}
Alfred Leick, Lev Rapoport, and Dmitry Tatarnikov.
\newblock {\em GPS satellite surveying}.
\newblock John Wiley \& Sons, 2015.

\bibitem{liu2017a}
Xinhua Liu, Huafeng Mei, Huachang Lu, Hailan Kuang, and Xiaolin Ma.
\newblock A {{Vehicle Steering Recognition System Based}} on {{Low}}-{{Cost
  Smartphone Sensors}}.
\newblock {\em Sensors}, 17(3):633, March 2017.

\bibitem{assessing_energy_trb1}
Qi~Luo, Joshua Auld, and Vadim Sokolov.
\newblock Addressing some issues of map-matching for large-scale,
  high-frequency gps data sets.
\newblock In {\em TRB 2015 Annual Meeting, Washington, DC}, 2015.

\bibitem{meiring2015}
Gys Albertus~Marthinus Meiring and Hermanus~Carel Myburgh.
\newblock A {{Review}} of {{Intelligent Driving Style Analysis Systems}} and
  {{Related Artificial Intelligence Algorithms}}.
\newblock {\em Sensors}, 15(12):30653--30682, December 2015.

\bibitem{nordfjaern2013}
Trond Nordfj{\ae}rn and Torbj{\o}rn Rundmo.
\newblock Road traffic safety beliefs and driver behaviors among personality
  subtypes of drivers in the {{Norwegian}} population.
\newblock {\em Traffic Injury Prevention}, 14(7):690--696, 2013.

\bibitem{ohn-bar2016}
E.~Ohn-Bar and M.~M. Trivedi.
\newblock Looking at {{Humans}} in the {{Age}} of {{Self}}-{{Driving}} and
  {{Highly Automated Vehicles}}.
\newblock {\em IEEE Transactions on Intelligent Vehicles}, 1(1):90--104, March
  2016.

\bibitem{paefgen2012}
Johannes Paefgen, Flavius Kehr, Yudan Zhai, and Florian Michahelles.
\newblock Driving {{Behavior Analysis}} with {{Smartphones}}: {{Insights}} from
  a {{Controlled Field Study}}.
\newblock In {\em Proceedings of the 11th {{International Conference}} on
  {{Mobile}} and {{Ubiquitous Multimedia}}}, MUM '12, pages 36:1--36:8, New
  York, NY, USA, 2012. {ACM}.

\bibitem{precht2017}
Lisa Precht, Andreas Keinath, and Josef~F. Krems.
\newblock Effects of driving anger on driver behavior \textendash{} {{Results}}
  from naturalistic driving data.
\newblock {\em Transportation Research Part F: Traffic Psychology and
  Behaviour}, 45:75--92, February 2017.

\bibitem{rudin1992nonlinear}
Leonid~I Rudin, Stanley Osher, and Emad Fatemi.
\newblock Nonlinear total variation based noise removal algorithms.
\newblock {\em Physica D: Nonlinear Phenomena}, 60(1-4):259--268, 1992.

\bibitem{sagberg2015}
Fridulv Sagberg, {Selpi}, Giulio~Francesco Bianchi~Piccinini, and Johan
  Engstr{\"o}m.
\newblock A {{Review}} of {{Research}} on {{Driving Styles}} and {{Road
  Safety}}.
\newblock {\em Human Factors}, 57(7):1248--1275, November 2015.

\bibitem{wallace2014}
B.~Wallace, R.~Goubran, F.~Knoefel, S.~Marshall, and M.~Porter.
\newblock Measuring variation in driving habits between drivers.
\newblock In {\em 2014 {{IEEE International Symposium}} on {{Medical
  Measurements}} and {{Applications}} ({{MeMeA}})}, pages 1--6, June 2014.

\bibitem{wallace2015}
B.~Wallace, A.~Puli, R.~Goubran, F.~Knoefel, S.~Marshall, M.~Porter, and
  A.~Smith.
\newblock Big data analytics to identify deceleration characteristics of an
  older driver.
\newblock In {\em 2015 {{IEEE International Symposium}} on {{Medical
  Measurements}} and {{Applications}} ({{MeMeA}}) {{Proceedings}}}, pages
  89--94, May 2015.

\bibitem{yu2016}
J.~Yu, H.~Zhu, H.~Han, Y.~J. Chen, J.~Yang, Y.~Zhu, Z.~Chen, G.~Xue, and M.~Li.
\newblock {{SenSpeed}}: {{Sensing Driving Conditions}} to {{Estimate Vehicle
  Speed}} in {{Urban Environments}}.
\newblock {\em IEEE Transactions on Mobile Computing}, 15(1):202--216, January
  2016.

\end{thebibliography}

\ifarxiv
\else
\begin{IEEEbiographynophoto}{Josh Warren}
Josh Warren is product manager at VideoAmp, a technology company that is building the TV operating system for advertising. Before that he was the director of analytics at Amnet - Denstu Aegis’ programmatic agency. There he developed bespoke data capabilities for advertisers including General Motors, Home Depot, Disney, Red Bull, Burberry, Discover, Norwegian Airlines, and Massage Envy. Warren holds a Master of Science in Analytics from the University of Chicago and a Bachelors of Science from Binghamton University in Economic Policy Analysis.Warren’s publications include healthcare delivery research presented in The Diabetes Educator, titled ‘The landscape for diabetes education: results of the 2012 AADE National Diabetes Education Practice Survey’.
\end{IEEEbiographynophoto}
\begin{IEEEbiographynophoto}{Jeff Lipkowitz}
Jeff Lipkowitz has worked in entrepreneurship, consulting and private equity . Jeff has focused on analytics and solution architecture throughout his career. He has championed several industries from health care to capital marketing. Helping transform the technology landscape for some of the worlds leading fortune100 companies. He has attended Purdue as well at University of Chicago focusing on operation research and analytics respectively. He currently works with startups by helping them in various capacities.
\end{IEEEbiographynophoto}

\begin{IEEEbiographynophoto}{Vadim Sokolov}
	Vadim Sokolov is an assistant professor in the Systems Engineering and Operations Research Department at George Mason University. He works on building robust solutions for large scale complex system analysis, at the interface of simulation-based modeling and statistics. This involves, developing new methodologies that rely on agent-based modeling, Bayesian analysis of time series data, design of computational experiments and development of open-source software that implements those methodologies. Inspired by an interest in urban
	systems he co-developed mobility simulator called Polaris that is
	currently used for large scale transportation networks analysis by
	both local and federal governments. Prior to joining Mason, he was a
	principal computational scientist at Argonne National Laboratory, a
	fellow at the Computation Institute at the University of Chicago, and
	lecturer at the Master of Science in Analytics program at the
	University of Chicago.
	
	Sokolov's publications include mathematics and engineering journals
	such as the Annals of Applied Statistics, Transportation Research Part
	C, Linear Algebra and Its Applications, as well as Mechanical Systems
	and Signal Processing. His recent work includes methods for sparse
	Bayesian estimation with application to high dimensional regression
	and classification.
\end{IEEEbiographynophoto}
\fi
\end{document}